\documentclass{article}

\usepackage[preprint]{neurips_2026}

\usepackage[utf8]{inputenc} 
\usepackage[T1]{fontenc}    
\usepackage{hyperref}       
\usepackage{url}            
\usepackage{booktabs}       
\usepackage{amsfonts}       
\usepackage{nicefrac}       
\usepackage{microtype}      
\usepackage{xcolor}         
\usepackage{adjustbox}
\usepackage{makecell}
\usepackage{tabularx}
\usepackage{float}
\usepackage{array}
\usepackage{placeins}
\usepackage{multirow} 
\usepackage{todonotes}
\usepackage{adjustbox}
\usepackage{enumitem}
\usepackage[ruled,vlined]{algorithm2e}

\title{Stashbird: Episode-Grounded Lifecycle Reconciliation for Conversational Agent Memory}

\author{
Chidera Biringa\thanks{Equal contribution. Correspondence to: \texttt{cbiringa@microsoft.com}.}
\quad
Lucas Yannul\footnotemark[1]
\quad
Xiaowen Wang
\quad
Marco Ayala
\\
\textbf{Nicholas Yi}
\quad
\textbf{Alex Moyse}
\quad
\textbf{Nishant Manchanda}
\quad
\textbf{Vivek Gupta}
\\
Microsoft
}

\begin{document}

\maketitle

\begin{abstract}

AI agents require memory that remains consistent as conversational evidence changes. We present Stashbird, an episode-grounded memory system that uses provenance to reconcile derived graph and retrieval state during updates and deletion. Repeated evidence extends support, corrections supersede prior active relations, and deletion removes unsupported state while preserving relations with remaining support. Under fixed semantic decisions, reconciliation produces correct final states in all 20 update sequences, compared with eight without reconciliation, and satisfies the defined checks across 263 deletion interventions. Across four long-term memory benchmarks, Stashbird achieves competitive question-answering accuracy and strong model efficiency. On LoCoMo, it uses 76.4x fewer ingestion prompt tokens than Graphiti and 8.1x fewer retrieval prompt tokens than reproduced Hindsight.

\end{abstract}

\section{Introduction}
\label{sec:introduction}
AI agents increasingly operate over interaction histories that cannot be handled reliably through full-context prompting alone~\cite{liu2024lost, hsiehruler, wulongmemeval, liu2023agentbench}. As these histories grow, especially in shared, multi-party settings, agents must retain facts, preferences, plans, and prior decisions while remaining responsive to corrections and changing evidence. A memory system for this scenario must recover earlier context, preserve information across interactions, track provenance, and support revision.

Current memory systems often attach an external memory layer to the model. Some rely on vector stores and retrieval-augmented generation~\cite{lewis2020retrieval, zhong2024memorybank}. Others use graph structures over entities, relations, and summaries~\cite{edge2024local, chhikara2025mem0, rasmussen2025zep}. Recent systems combine dense retrieval with structured or hierarchical memory~\cite{latimer2025hindsight, tang2026mnemis}. These systems must balance answer quality against the model calls and tokens required to construct and query memory.

Memory must also remain consistent as conversational evidence changes~\cite{lam2026governing}. Recent work studies this problem through forgetting-aware evaluation, selective graph pruning, and the removal of information from derived agent state~\cite{uddin2026recall,rusu2026selective,yao2026forgetting}. Existing memory systems also provide episodic provenance, contradiction handling, and update operations~\cite{chhikara2025mem0,rasmussen2025zep,li2025memos}. Stashbird addresses how graph and search records should be updated when a source episode is added, corrected, or removed. It preserves facts that retain evidence support, withdraws unsupported records, maintains graph connectivity, and removes affected community summaries from search until they are regenerated.

We present Stashbird, an agent memory system that links source episodes to the entities and relations derived from them. These links allow the system to update memory when evidence changes. Repeated evidence adds support to an existing relation. Corrections replace the active relation while retaining its earlier version as history. When an episode is deleted, Stashbird removes records supported only by that episode and preserves records supported by other episodes. It also clears metadata from unsupported entities needed to keep the graph connected and removes affected community summaries from search until they are regenerated.

We evaluate Stashbird through static question-answering benchmarks and separate lifecycle studies. The benchmark evaluation measures accuracy and model-facing workload across four long-term memory benchmarks. The lifecycle evaluation measures final-state correctness under controlled update and deletion interventions.

Our contributions are as follows:
\begin{itemize}
\item We introduce evidence-support lifecycle reconciliation for structured conversational memory. Provenance determines how derived relations and retrieval-visible state are extended, superseded, preserved, or removed as source episodes change.

\item We present an episode-level deletion procedure that preserves multiply supported relations, removes unsupported records, retains connective graph structure without stale metadata, invalidates affected community documents, cleans the retrieval index, and repairs the episode chain.

\item We evaluate Stashbird through static question-answering and lifecycle experiments. The lifecycle studies isolate reconciliation under fixed semantic decisions, while four long-term memory benchmarks measure answer accuracy and model efficiency.
\end{itemize}

\begin{figure*}[ht]
\centering
\includegraphics[width=0.90\textwidth]{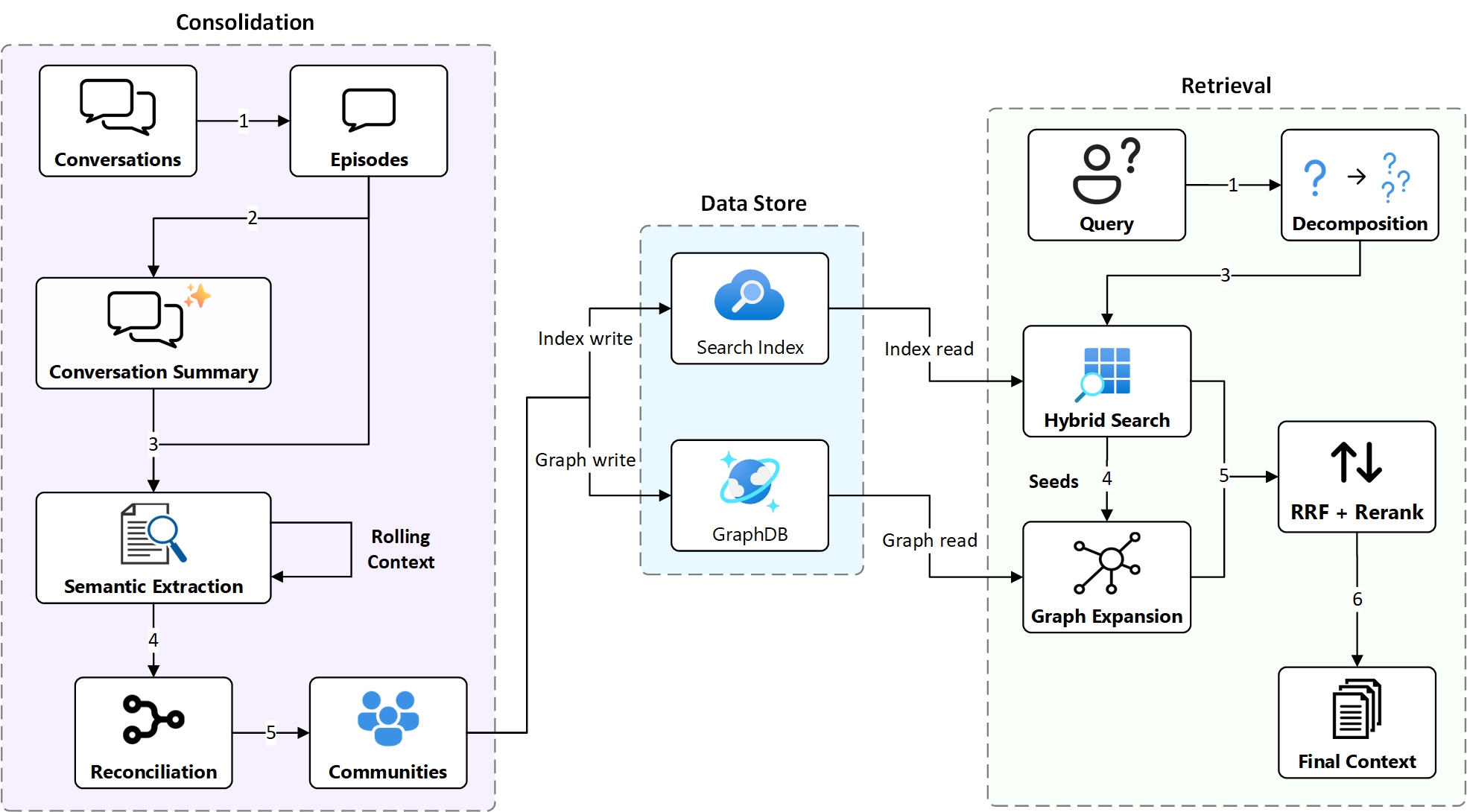}
\caption{Overview of Stashbird. The purple region shows consolidation, where conversations are converted into episodes, summaries, semantic memories, and communities. The blue region shows persistent storage in the search index and graph database. The green region shows retrieval over the stored memory using query decomposition, hybrid search, rank fusion, and reranking.} 
\label{fig:stashbird_overview}
\end{figure*}


\begin{table}[ht]
\centering
\caption{Structured memory views in Stashbird and their roles.}
\label{tab:memory-views}

\adjustbox{max width=\columnwidth}{%
\begin{tabular}{llll}
\toprule
View & Stored Unit & Update Behavior & Used For \\
\midrule
Episodes & Conversation chunks & Append / delete & Evidence recall \\
Entities & Canonical entities & Merge / invalidate & Grounding \\
Relations & Entity relations & Replace / version & Reasoning \\
Preferences & User traits & Overwrite & Personalization \\
Communities & Entity clusters & Re-summarize & High-level context \\
\bottomrule
\end{tabular}
}
\end{table}

\section{Stashbird}
\label{sec:vulStyle}

Figure~\ref{fig:stashbird_overview} shows ingestion and update (\S~\ref{subsection:ingestion}), deletion (\S~\ref{subsection:deletion}), and retrieval (\S~\ref{subsection:retrieval}). Ingestion creates ordered conversation episodes and derives entities and relations from them. Update appends new episodes to persisted memory. Both operations retain the source episode identifiers associated with derived entities and relations. Repeated evidence adds support to existing relations, while corrections supersede earlier active relations. Deletion uses the same source links to remove unsupported state, preserve state supported by other episodes, and mark affected communities for re-summarization. Retrieval selects relevant evidence from the resulting memory for question answering. The static benchmarks evaluate ingestion and retrieval, while separate lifecycle experiments evaluate update and deletion.

\subsection{Memory Ingestion and Update}
\label{subsection:ingestion}

Stashbird ingests conversations as ordered timestamped episodes and constructs a persistent memory state that can be searched, updated, and selectively removed. Let \( E = \{e_1, \dots, e_T\} \) denote the episode sequence, where each \( e_t \) stores the turn author, content, message time, and ingestion time. Adjacent episodes form a conversational thread, and episodes link to referenced entities while preserving provenance on entity and relation records. To avoid repeated full-history LLM calls, Stashbird first summarizes the conversation and then processes episodes in overlapping chunks. A named-entity recognition model~\cite{honnibal2020spacy} reduces missed mentions before the backbone extracts entities and relations from the chunk text, global summary, and rolling entity-centered context.

The rolling context stores recent summaries and relation facts within a fixed token budget. It is compressed when needed and then pruned by removing the least-supported entities if compression remains insufficient. Extracted entities and relations are reconciled against the existing state. Entity reconciliation uses both embedding similarity and name similarity. A candidate is accepted automatically only when both signals strongly agree. Candidates supported by only one signal are passed to the backbone for resolution. This limits automatic merging to candidates supported by independent lexical and semantic evidence. Relations are reconciled within each entity pair, where duplicates extend provenance and contradictions replace the earlier relation.

Stashbird exposes memory through structured views over the same conversation state, including episodes, entities, relations, preference-like facts, and community summaries. Searchable documents support semantic and lexical retrieval over these views. Table~\ref{tab:memory-views} summarizes the system's memory views. Dedicated preference retrieval was disabled in the reported experiments. When community summarization is enabled, Stashbird clusters the entity graph with label propagation~\cite{raghavan2007label}. Each cluster is summarized and stored as a community node linked to its members.

\noindent \textbf{Update. }
Stashbird supports incremental updates by appending new episodes to an existing memory graph. A persisted checkpoint stores the conversation summary, serialized rolling context, last processed episode, and current entity count. During updates, the system loads the checkpoint, restores the rolling context, and appends new episodes to the conversational thread. The conversation summary is updated using the added episodes. Semantic consolidation processes the appended turns and reuses existing entities and relations. When communities are enabled, newly created entities are assigned to existing communities

\begin{algorithm}[ht]
\small
\LinesNumbered
\DontPrintSemicolon
\caption{Episode-Anchored Deletion}
\label{alg:deletion}
\KwIn{Episode \(d\), memory state \(\mathcal{M}\)}
\KwOut{Updated memory state \(\mathcal{M}\)}

\(\mathcal{E} \gets \textsc{AffectedEntities}(d, \mathcal{M})\)\;
\(\mathcal{R} \gets \textsc{AffectedRelations}(d, \mathcal{M})\)\;
\(\mathcal{C} \gets \textsc{AffectedCommunities}(\mathcal{E}, \mathcal{M})\)\;
\((p,n) \gets \textsc{EpisodeNeighbors}(d, \mathcal{M})\)\;

\ForEach{\(r \in \mathcal{R}\)}{
    \(P(r) \gets P(r) \setminus \{d\}\)\;
    \If{\(P(r) = \emptyset\)}{
        Remove \(r\) from the graph and search index\;
    }
    \Else{
        Update provenance for \(r\)\;
    }
}

\ForEach{\(e \in \mathcal{E}\)}{
    \(P(e) \gets P(e) \setminus \{d\}\)\;
    \If{\(P(e) \neq \emptyset\)}{
        Update provenance for \(e\)\;
    }
    \ElseIf{\(e\) has no surviving neighbors}{
        Remove \(e\) from the graph and search index\;
    }
    \Else{
        Clear episode-derived metadata for \(e\)\;
        Preserve \(e\) as a connective anchor\;
        Remove its stale search document\;
    }
}

Remove \(d\) from the graph and search index\;

\If{\(p\) and \(n\) exist}{
    Link \(p\) to \(n\)\;
}

Mark communities in \(\mathcal{C}\) dirty and remove their stale search documents\;

\Return{\(\mathcal{M}\)}
\end{algorithm}

\subsection{Memory Deletion}
\label{subsection:deletion}
Stashbird performs deletion at the episode level. A deletion request specifies an episode identifier, and provenance~\cite{cheney2009provenance} identifies the entities, relations, and communities affected by that episode. Let \(P(x)\) denote the set of episodes supporting entity or relation \(x\), and let \(d\) be the episode to delete. Algorithm~\ref{alg:deletion} summarizes the complete reconciliation process. For each affected relation \(r\), Stashbird updates \(P(r) \leftarrow P(r) \setminus \{d\}\). Relations with remaining support are retained with updated provenance. Relations without support are removed from the graph and search index.

For each affected entity \(e\), Stashbird updates \(P(e) \leftarrow P(e) \setminus \{d\}\) and applies Entity Meta-forgetting with Neighbor Adoption. Entities with remaining support receive only provenance updates. Unsupported entities without surviving neighbors are removed. Unsupported entities with surviving neighbors retain their structural identity as connective anchors after their episode-derived metadata is cleared~\cite{hogan2021knowledge}. Removed entities and stale entity documents are also removed from the search index. Finally, Stashbird removes the episode from the graph and search index, reconnects its predecessor and successor when both exist, and marks affected communities dirty for deferred re-summarization. Stale community documents are removed from retrieval until their summaries are regenerated.

\subsection{Memory Retrieval}
\label{subsection:retrieval}
Stashbird retrieves evidence from its episode-linked memory state. We retain the original query and use the backbone to generate up to four focused sub-queries for questions involving multiple entities, time constraints, or related facts. 

Let \(q\) denote the input query and \(Q = \{q_0, \dots, q_k\}\) the set of query variants, where \(q_0\) is the original query. Each query variant is embedded, and temporal expressions are resolved relative to the question date when available. Conversation-scoped hybrid search retrieves episode, entity, and relation documents, together with community documents where enabled. Results across queries are merged and deduplicated. Evidence from different sources is fused using reciprocal rank fusion~\cite{cormack2009reciprocal}. An LLM relevance scorer assigns each fused candidate a score based on its usefulness for answering the query. The top-ranked candidates form the final evidence set. The response generator passes this context, together with the question date, to the backbone, which is instructed to answer using only the retrieved evidence.

\section{Experiments}
\label{sec:experiments}

\begin{table}[ht]
\centering
\small
\caption{LLM-as-a-judge accuracy on LongMemEval-S by question type.
SS-User: Single-Session User,
MS: Multi-Session,
SS-Pref: Single-Session Preference,
Temp: Temporal Reasoning,
KU: Knowledge Update,
Full Con: Full Context,
SS-Asst: Single-Session Assistant.
For reproduced methods with repeated runs, Overall reports the mean $\pm$ 95\% confidence interval over 5 runs. Results above the separator are taken directly from prior reports, while those below are reproduced in our experiments. Bold indicates the highest score in each column within each backbone group.}
\setlength{\tabcolsep}{3.5pt}
\adjustbox{max width=\columnwidth}{
\begin{tabular}{p{1.8cm} p{1.2cm} c c c c c c c}
\toprule
\textbf{Backbone} & \textbf{Method}
& SS-User
& MS
& SS-Pref
& Temp
& KU
& SS-Asst
& \textbf{Overall} \\
\midrule

\multirow{6}{*}{GPT-4o-mini}
& Zep & 92.9 & 47.4 & 53.3 & 54.1 & 74.4 & 75.0 & 63.2 \\
& Nemori & 88.6 & 51.1 & 46.7 & 61.7 & 61.5 & 83.9 & 64.2 \\
\cmidrule(lr){2-9}
& Full Con & 84.6 & 44.1 & 9.5 & 45.4 & \textbf{81.0} & 93.5 & 59.8 \\
& Mem0 & 90.6 & 53.5 & 39.3 & 59.5 & 75.2 & 30.4 & $59.8{\scriptstyle \pm 0.4}$ \\
& Hindsight & 93.1 & \textbf{67.1} & 53.7 & \textbf{68.0} & 75.7 & 95.7 & $74.8{\scriptstyle \pm 1.8}$ \\
& \textbf{Stashbird} & \textbf{95.6} & 63.2 & \textbf{74.6} & 67.1 & 77.5 & \textbf{97.1} & $\mathbf{75.6}{\scriptstyle \pm 0.9}$ \\

\midrule

\multirow{5}{*}{GPT-4.1-mini}
& Nemori & 90.0 & 55.6 & 86.7 & 72.2 & 79.5 & 92.9 & 74.6 \\
\cmidrule(lr){2-9}
& Full Con & 94.3 & 53.6 & 10.6 & 60.4 & 78.2 & \textbf{98.5} & 67.4 \\
& Mem0 & 95.9 & \textbf{70.3} & 42.9 & \textbf{77.5} & 78.1 & 26.4 & $70.0{\scriptstyle \pm 0.6}$ \\
& Hindsight & 94.7 & 67.2 & 55.3 & 76.5 & 81.1 & 96.4 & $78.3{\scriptstyle \pm 2.6}$ \\
& \textbf{Stashbird} & \textbf{98.6} & 69.9 & \textbf{88.4} & 72.4 & \textbf{84.0} & 95.9 & $\mathbf{81.0}{\scriptstyle \pm 0.8}$ \\

\bottomrule
\end{tabular}}
\label{tab:longmemeval}
\end{table}

\begin{table}[ht]
\centering
\small
\caption{LLM-as-a-judge accuracy on LoCoMo by question type. Following prior work, the Adversarial category is excluded. For reproduced methods with repeated runs, Overall reports the mean $\pm$ 95\% confidence interval over five runs. MH: Multi Hop, Temp: Temporal, OD: Open Domain, Full Con: Full Context, SH: Single Hop}
\setlength{\tabcolsep}{3.5pt}
\adjustbox{max width=\columnwidth}{
\begin{tabular}{p{1.8cm} p{1.4cm} c c c c c}
\toprule
\textbf{Backbone} & \textbf{Method}
& \makecell{MH}
& \makecell{Temp}
& \makecell{OD}
& \makecell{SH}
& \textbf{Overall} \\
\midrule

\multirow{7}{*}{GPT-4o-mini}
& Zep & 50.5 & 58.9 & 39.6 & 63.2 & 58.5 \\
& Nemori & 65.3 & 71.0 & 44.8 & 82.1 & 74.4 \\
\cmidrule(lr){2-7}
& Full Con & 79.5 & 57.9 & \textbf{74.4} & \textbf{91.3} & 80.9 \\
& Mem0 & 74.0 & 63.3 & 60.0 & 74.9 & $71.4{\scriptstyle \pm 0.3}$ \\
& Graphiti & 61.5 & 46.2 & 57.7 & 58.3 & $56.3{\scriptstyle \pm 0.6}$ \\
& Hindsight & \textbf{87.5} & \textbf{77.7} & 73.2 & 89.9 & $\mathbf{85.9}{\scriptstyle \pm 0.4}$ \\
& \textbf{Stashbird} & 79.0 & 69.3 & 57.9 & 87.7 & $80.4{\scriptstyle \pm 0.7}$ \\

\midrule

\multirow{7}{*}{GPT-4.1-mini}
& Zep & 53.7 & 60.2 & 43.8 & 66.9 & 61.6 \\
& Nemori & 75.1 & 77.6 & 51.0 & 84.9 & 79.5 \\
\cmidrule(lr){2-7}
& Full Con & 81.5 & 78.8 & 62.5 & 88.9 & 83.8 \\
& Mem0 & 82.4 & 72.1 & 66.7 & 78.3 & $77.0{\scriptstyle \pm 0.4}$ \\
& Graphiti & 68.4 & 67.9 & 65.0 & 74.3 & $71.3{\scriptstyle \pm 0.4}$ \\
& Hindsight & \textbf{88.0} & \textbf{86.3} & \textbf{74.7} & \textbf{92.2} & $\mathbf{89.1}{\scriptstyle \pm 0.7}$ \\
& \textbf{Stashbird} & 87.4 & 83.5 & 68.5 & 91.3 & $87.5{\scriptstyle \pm 0.3}$ \\

\bottomrule
\end{tabular}}
\label{tab:locomo}
\end{table}

\subsection{Datasets}
We evaluate Stashbird on four long-term memory benchmarks: LongMemEval-S\cite{wulongmemeval}, LoCoMo\cite{maharana2024evaluating}, EverMemBench\cite{hu2026evermembench}, and GroupMemBench\cite{groupmembench}. LongMemEval-S tests user-assistant memory over long histories, with 500 questions and roughly 115K tokens per question, covering information extraction, multi-session reasoning, temporal reasoning, knowledge updates, and abstention. LoCoMo evaluates dyadic multi-session conversations with single-hop, multi-hop, temporal, and open-domain questions. 

EverMemBench and GroupMemBench extend evaluation to shared settings: EverMemBench uses multi-party, multi-group dialogues for fine-grained recall, memory awareness, and user profile understanding, while GroupMemBench uses role-tagged enterprise group-channel logs across Finance, Technology, Healthcare, and Manufacturing, covering multi-hop reasoning, updates, temporal reasoning, user-conditioned queries, term ambiguity, and abstention. These shared-conversation benchmarks are central to Stashbird because memory must persist across interactions where users may communicate with or without an agent present.

\subsection{Implementation Details}
We use Azure OpenAI in Azure AI Foundry for all LLM and embedding calls. Stashbird uses 3072-dimensional \texttt{text-embedding-3-large} embeddings and the \texttt{gpt-4o-mini} (\texttt{2024-07-18}), \texttt{gpt-4.1-mini} (\texttt{2025-04-14}), \texttt{gpt-5.2-chat} (\texttt{2026-02-10}), and \texttt{gpt-5.4} (\texttt{2026-03-05}) backbones. The sensitivity analyses vary the backbone and final evidence count. We fix all other settings across benchmarks.

Stashbird processes conversations in 20K-token chunks with 10\% overlap, using a 15K-token summary window and a 4K-token rolling context. These limits bound each model call and repeated text across adjacent chunks. Entity reconciliation retrieves five vector candidates. Stashbird accepts a candidate when cosine similarity and Levenshtein ratio both exceed 0.85 and sends candidates exceeding 0.75 on either measure to the backbone. This design requires strong lexical and semantic agreement for automatic merges while retaining more candidates for model-based resolution.

Retrieval uses top-60 hybrid search per document type, 30 rerank candidates, and 30 final evidence items. RRF uses the standard \(k=60\) setting~\cite{cormack2009reciprocal}. Table~\ref{tab:sensitivity} evaluates final evidence counts of 10, 20, 30, and 50. We disable graph expansion, secondary coverage expansion, credibility weighting, reflection, and dedicated preference retrieval, so their settings do not affect the results.

\subsection{Performance Evaluation}
\noindent \textbf{Baselines.}
We compare Stashbird with Full Context, Zep~\cite{rasmussen2025zep}, Mem0~\cite{chhikara2025mem0}, Nemori~\cite{nan2025nemori}, MemOS~\cite{li2025memos}, Hindsight~\cite{latimer2025hindsight}, MemoBase~\cite{memobase}, and Graphiti~\cite{graphiti}. We reproduce Full Context, Mem0, Graphiti, Hindsight, and Stashbird with a common harness using the same questions, answer backbone, and judge configuration within each benchmark. Each system retains its native embedding and retrieval configuration. We report the remaining baselines from prior work, using Supermemory for LongMemEval-S, Backboard LoCoMo for LoCoMo, and the EverMemBench paper for EverMemBench. The tables separate reported and reproduced results.

\noindent \textbf{Hardware and Criterion.}
We run one benchmark per Azure Batch \texttt{Standard\_E32ds\_v4} node with 32 vCPUs, 252~GiB RAM, an Intel Xeon Platinum 8272CL CPU, and Ubuntu 22.04. We follow each benchmark's LLM-as-a-judge protocol. LongMemEval-S uses binary yes/no correctness, and LoCoMo uses Correct/Wrong labels with the adversarial category excluded. EverMemBench uses its dataset answer generator and judge. GroupMemBench uses its benchmark judge and a separate rubric for abstention questions.

\noindent \textbf{Model-facing workload.}
We measure LLM calls, embedding calls, prompt tokens, and completion tokens during ingestion and retrieval. The counts cover extraction, entity resolution, enabled community summarization, query decomposition, reranking, and answer generation. We report dataset-wide totals averaged over repeated iterations.

\section{Results}
\label{sec:results}

\begin{table}[t]
\centering
\small
\caption{GroupMemBench accuracy with GPT-5.2-chat on the released Finance, Technology, Healthcare, and Manufacturing sets. MH: Multi-hop; KU: Knowledge Update; Temp: Temporal; UI: User Implicit; TA: Term Ambiguity; Abst: Abstention; Avg: mean across query types.}
\label{tab:groupmembench}
\setlength{\tabcolsep}{3.5pt}
\adjustbox{max width=\columnwidth}{
\begin{tabular}{lccccccc}
\toprule
\textbf{Method} & \textbf{MH} & \textbf{KU} & \textbf{Temp} & \textbf{UI} & \textbf{TA} & \textbf{Abst} & \textbf{Avg} \\
\midrule
Hindsight & 18.9 & 13.3 & 2.2 & 22.8 & 28.0 & \textbf{64.5} & $24.9{\scriptstyle \pm 1.0}$ \\
\textbf{Stashbird} & \textbf{52.9} & \textbf{35.8} & \textbf{72.3} & \textbf{60.8} & \textbf{29.1} & 55.7 & $\mathbf{51.1}{\scriptstyle \pm 1.9}$ \\
\bottomrule
\end{tabular}}
\end{table}



\begin{table}[ht]
\centering
\caption{Ingestion and retrieval model usage on LongMemEval-S, LoCoMo, and EverMemBench using GPT-4.1-mini. Ingestion values are one-time totals. Retrieval values are means over repeated retrieval iterations. Counts are abbreviated with K, M, and B for thousands, millions, and billions.}

\setlength{\tabcolsep}{3pt}
\renewcommand{\arraystretch}{0.95}
\adjustbox{max width=\columnwidth}{
\begin{tabular}{l l p{2.0cm} c c c c}
\toprule
\textbf{Benchmark} & \textbf{Phase} & \textbf{Method}
& \makecell{LLM \\ Calls}
& \makecell{Embedding \\ Calls}
& \makecell{Prompt \\ Tokens}
& \makecell{Completion \\ Tokens} \\
\midrule

\multirow{8}{*}{LongMemEval-S}
& \multirow{4}{*}{Ingestion}
& Mem0 & 109K & 291K & 1.03B & \textbf{8.98M} \\
& & Graphiti & 1.28M & 2.07M & 3.57B & 108M \\
& & Hindsight & 137K & 462K & 700M & 81.6M \\
& & \textbf{Stashbird} & \textbf{29K} & \textbf{29K} & \textbf{164M} & 13M \\

\cmidrule(lr){2-7}

& \multirow{4}{*}{Retrieval}
& Mem0 & \textbf{470} & 1.09K & 6.75M & 76.9K \\
& & Graphiti & 9.84K & \textbf{940} & \textbf{2.04M} & \textbf{23.8K} \\
& & Hindsight & 1.49K & 1.14K & 19.9M & 139K \\
& & \textbf{Stashbird} & 3.49K & 2.47K & 5.51M & 76.9K \\

\midrule

\multirow{8}{*}{LoCoMo}
& \multirow{4}{*}{Ingestion}
& Mem0 & 5.08K & 14.6K & 44.6M & 414K \\
& & Graphiti & 18K & 27K & 37M & 742K \\
& & Hindsight & 796 & 4.48K & 6.58M & 507K \\
& & \textbf{Stashbird} & \textbf{57} & \textbf{294} & \textbf{484K} & \textbf{34.0K} \\

\cmidrule(lr){2-7}

& \multirow{4}{*}{Retrieval}
& Mem0 & \textbf{1.54K} & 4.13K & 19.7M & 932K \\
& & Graphiti & 32.2K & \textbf{3.08K} & \textbf{6.87M} & \textbf{131K} \\
& & Hindsight & 4.92K & 3.20K & 66.4M & 460K \\
& & \textbf{Stashbird} & 10.8K & 7.65K & 8.24M & 216K \\

\midrule

\multirow{4}{*}{EverMemBench}
& \multirow{2}{*}{Ingestion}
& Hindsight & 12.2K & 71.8K & 169M & 12.0M \\
& & \textbf{Stashbird} & \textbf{3.97K} & \textbf{2.87K} & \textbf{21.4M} & \textbf{1.57M} \\

\cmidrule(lr){2-7}

& \multirow{2}{*}{Retrieval}
& Hindsight & \textbf{12.1K} & \textbf{8.57K} & 282M & 1.84M \\
& & \textbf{Stashbird} & 37.4K & 15.6K & \textbf{52.5M} & \textbf{1.12M} \\

\bottomrule
\end{tabular}}
\label{tab:efficiency}
\end{table}

\begin{table*}[t]
\centering
\small
\caption{Sensitivity to final evidence count and answer backbone. \(K\) is the number of evidence items passed to the answer model. The \(K\) rows use GPT-4.1-mini, and the backbone rows use \(K=30\). Avg. is the EverMemBench mean; Overall is LoCoMo accuracy.}
\label{tab:sensitivity}
\setlength{\tabcolsep}{2.2pt}
\adjustbox{max width=\textwidth}{
\begin{tabular}{l c c c c c c c c c c c c c c c}
\toprule
\textbf{Setting}
& \multicolumn{10}{c}{\textbf{EverMemBench}}
& \multicolumn{5}{c}{\textbf{LoCoMo}} \\
\cmidrule(lr){2-11}\cmidrule(lr){12-16}
& Single & Multi & Temp & Const & Proact & Update & Style & Skill & Role & \textbf{Avg.}
& \makecell{Multi-\\Hop} & Temporal & \makecell{Open\\Domain}
& \makecell{Single\\Hop} & \textbf{Overall} \\
\midrule
\(K=10\)
& 95.7 & 19.4 & \textbf{23.1} & 84.2 & 58.8 & 61.3 & 41.9 & 50.9 & 61.3 & 55.2
& \textbf{89.0} & 81.7 & 66.7 & 90.9 & 87.1 \\
\(K=20\)
& \textbf{95.8} & 21.6 & 19.3 & 84.2 & \textbf{59.4} & 63.6 & 42.8 & 50.6 & \textbf{62.2} & \textbf{55.5}
& 88.8 & 81.4 & 68.3 & 90.6 & 87.0 \\
\(K=30\)
& 95.6 & 20.3 & 17.9 & \textbf{84.4} & 58.8 & 66.7 & \textbf{43.3} & 51.0 & 60.9 & 55.4
& 87.4 & \textbf{83.5} & \textbf{68.5} & \textbf{91.3} & \textbf{87.5} \\
\(K=50\)
& 95.7 & \textbf{23.5} & 16.9 & 83.9 & 58.0 & \textbf{67.6} & 40.2 & \textbf{52.1} & 60.1 & 55.4
& 88.7 & 82.1 & 66.0 & 91.0 & 87.1 \\
\midrule
GPT-4.1-mini
& 95.6 & 20.3 & 17.9 & 84.4 & 58.8 & 66.7 & \textbf{43.3} & 51.0 & \textbf{60.9} & 55.4
& 87.4 & 83.5 & 68.5 & 91.3 & 87.5 \\
GPT-5.2-chat
& 95.7 & \textbf{20.7} & 19.1 & \textbf{86.0} & \textbf{63.5} & \textbf{73.4} & 39.8 & \textbf{52.2} & 57.2 & \textbf{56.4}
& 88.4 & 86.8 & 77.9 & 93.0 & 89.9 \\
GPT-5.4
& \textbf{96.6} & 20.2 & \textbf{24.9} & 85.4 & 61.3 & 69.7 & 39.7 & 49.7 & 55.7 & 55.9
& \textbf{91.9} & \textbf{91.0} & \textbf{78.5} & \textbf{93.8} & \textbf{92.0} \\
\bottomrule
\end{tabular}}
\end{table*}

\subsection{Benchmark Performance}

\noindent \textbf{LongMemEval-S.}
We report LongMemEval-S accuracy by question type in Table~\ref{tab:longmemeval}. Stashbird scores 75.6\% overall with GPT-4o-mini, slightly above Hindsight at 74.8\%, and 81.0\% with GPT-4.1-mini, 2.7 points above Hindsight at 78.3\%. It also outperforms Zep, Nemori, Full Context, and Mem0. 

The largest difference is on single-session preference questions, where Stashbird scores 74.6\% and 88.4\% across the two backbones, compared with 53.7\% and 55.3\% for Hindsight. Stashbird also shows lower run-to-run variance than Hindsight, with narrower confidence intervals across both backbones.

\begin{table}[ht]
\centering
\small
\caption{LLM-as-a-judge accuracy on EverMemBench across three categories using GPT-4.1-mini as the backbone for all methods:
Fine-Grained Recall (Single: Single-Hop, Multi: Multi-Hop, Temp: Temporal),
Memory Awareness (Const: Constraint, Proact: Proactivity, Update),
and Profile Understanding (Style, Skill, Role). We report mean accuracy percentages. Avg is the unweighted mean across the nine sub-tasks. For the methods we evaluate, it is shown as mean $\pm$ 95\% confidence interval over five runs. Results above the separator are taken directly from prior reports, while those below are reproduced in our experiments.} 
\setlength{\tabcolsep}{2.5pt}
\adjustbox{max width=\columnwidth}{
\begin{tabular}{p{1.5cm} ccc ccc ccc c}
\toprule
\textbf{Method}
& \multicolumn{3}{c}{Fine-Grained Recall}
& \multicolumn{3}{c}{Memory Awareness}
& \multicolumn{3}{c}{Profile Understanding}
& \textbf{Avg} \\
\cmidrule(lr){2-4} \cmidrule(lr){5-7} \cmidrule(lr){8-10}
& Single & Multi & Temp & Const & Proact & Update & Style & Skill & Role & \\
\midrule

MemoBase & 60.0 & 12.8 & 18.0 & 64.6 & 36.7 & 30.6 & 17.0 & 29.5 & 38.7 & 34.2 \\
Zep & 73.7 & 8.0 & 13.0 & 67.1 & 47.5 & 43.6 & 26.7 & 35.5 & 44.3 & 39.9 \\
MemOS & 71.3 & 18.8 & 15.6 & 69.9 & 51.9 & 45.1 & 28.9 & 32.5 & 48.4 & 42.5 \\
\cmidrule(lr){1-11}
Hindsight & 90.8 & \textbf{22.8} & \textbf{22.4} & 82.3 & \textbf{66.9} & \textbf{68.6} & 33.5 & 49.9 & 56.5 & $54.8{\scriptstyle \pm 0.5}$ \\
\textbf{Stashbird} & \textbf{95.6} & 20.3 & 17.9 & \textbf{84.4} & 58.8 & 66.7 & \textbf{43.3} & \textbf{51.0} & \textbf{60.9} & $\mathbf{55.4}{\scriptstyle \pm 0.5}$ \\

\bottomrule
\end{tabular}}
\label{tab:evermembench}
\end{table}

\noindent \textbf{LoCoMo.}
Table~\ref{tab:locomo} presents LoCoMo accuracy by question type. With GPT-4o-mini, Stashbird achieves 80.4\% overall, outperforming Zep, Nemori, Mem0, and Graphiti and nearly matching Full Context at 80.9\%. With GPT-4.1-mini, Stashbird improves to 87.5\%, outperforming Zep, Nemori, Full Context, Mem0, and Graphiti. Hindsight remains higher at 85.9\% and 89.1\%, but Section~\ref{subsection:efficiency} shows that this accuracy gap comes with substantially higher cost.

\noindent \textbf{EverMemBench.}
For EverMemBench, Table~\ref{tab:evermembench} shows that Stashbird achieves 55.4\% average accuracy with GPT-4.1-mini, outperforming MemoBase, Zep, and MemOS, and performing comparably to reproduced Hindsight at 54.8\%. Stashbird is strongest on single-hop recall, constraint awareness, and profile-understanding questions, while Hindsight performs better on multi-hop recall, temporal recall, proactivity, and knowledge update. This pattern suggests that Stashbird performs best when answers rely on explicit user or group evidence, while broader evidence chaining and behavioral inference remain challenging.

\noindent \textbf{GroupMemBench.}
On GroupMemBench, Table~\ref{tab:groupmembench} shows that Stashbird achieves 51.1\% average accuracy, 26.2 points above Hindsight at 24.9\%. The largest differences are on temporal reasoning, 72.3\% versus 2.2\%, and user-implicit questions, 60.8\% versus 22.8\%. The strong temporal result is consistent with Stashbird preserving speaker and time information across shared conversation.

\subsection{Model Usage}
\label{subsection:efficiency}
Table~\ref{tab:efficiency} reports ingestion and retrieval model calls and token usage across benchmarks.

\noindent \textbf{LongMemEval-S.}
Stashbird uses 29K ingestion LLM calls, compared with 109K for Mem0, 137K for Hindsight, and 1.3M for Graphiti. It also uses 164M ingestion prompt tokens, compared with 700M for Hindsight and 3.6B for Graphiti. At retrieval time, Stashbird uses more LLM calls than Mem0 and Hindsight but fewer prompt tokens, 5.5M compared with 19.9M for Hindsight and 6.8M for Mem0.

\noindent \textbf{LoCoMo.}
Stashbird uses 57 ingestion LLM calls, compared with 5.08K for Mem0, 18K for Graphiti, and 796 for Hindsight, and 484K ingestion prompt tokens, compared with 44.6M, 37M, and 6.58M. At retrieval time, Stashbird uses more LLM calls than Mem0 and Hindsight, but fewer prompt tokens, 8.24M compared with 19.7M for Mem0 and 66.4M for Hindsight.

\noindent \textbf{EverMemBench.}
Stashbird uses 3.97K ingestion LLM calls and 21.4M ingestion prompt tokens, compared with 12.2K calls and 169M prompt tokens for Hindsight. At retrieval time, Stashbird uses more LLM calls than Hindsight but fewer prompt tokens, 52.5M compared with 282M.

\subsection{Ablation and Sensitivity Studies}

\noindent \textbf{Backbone Sensitivity.}
Table~\ref{tab:sensitivity} shows that stronger backbones improve Stashbird consistently on LoCoMo, where overall accuracy rises from 87.5\% with GPT-4.1-mini to 89.9\% with GPT-5.2-chat and 92.0\% with GPT-5.4. The largest gains occur on open-domain and temporal questions, which require synthesizing retrieved evidence rather than extracting a short fact. On EverMemBench, scaling has a smaller and non-monotonic effect: average accuracy changes from 55.4\% to 56.4\% and 55.9\%, with gains in temporal recall and knowledge update but flat or lower profile-understanding scores. This suggests that EverMemBench is more retrieval-bounded than backbone-bounded.

\noindent \textbf{Influence of Top K.}
Table~\ref{tab:sensitivity} varies the number of retrieved evidence items passed to the answer model. Across $K \in \{10,20,30,50\}$, performance remains stable. EverMemBench average accuracy stays between 55.2\% and 55.5\%, while LoCoMo overall accuracy stays between 87.0\% and 87.5\%, with the best score at $K=30$. Increasing $K$ therefore adds candidates but does not reliably improve answers, suggesting that useful evidence is usually already present in the top-ranked context after fusion and reranking.

\begin{figure}[ht]
\centering
\includegraphics[width=0.80\textwidth]{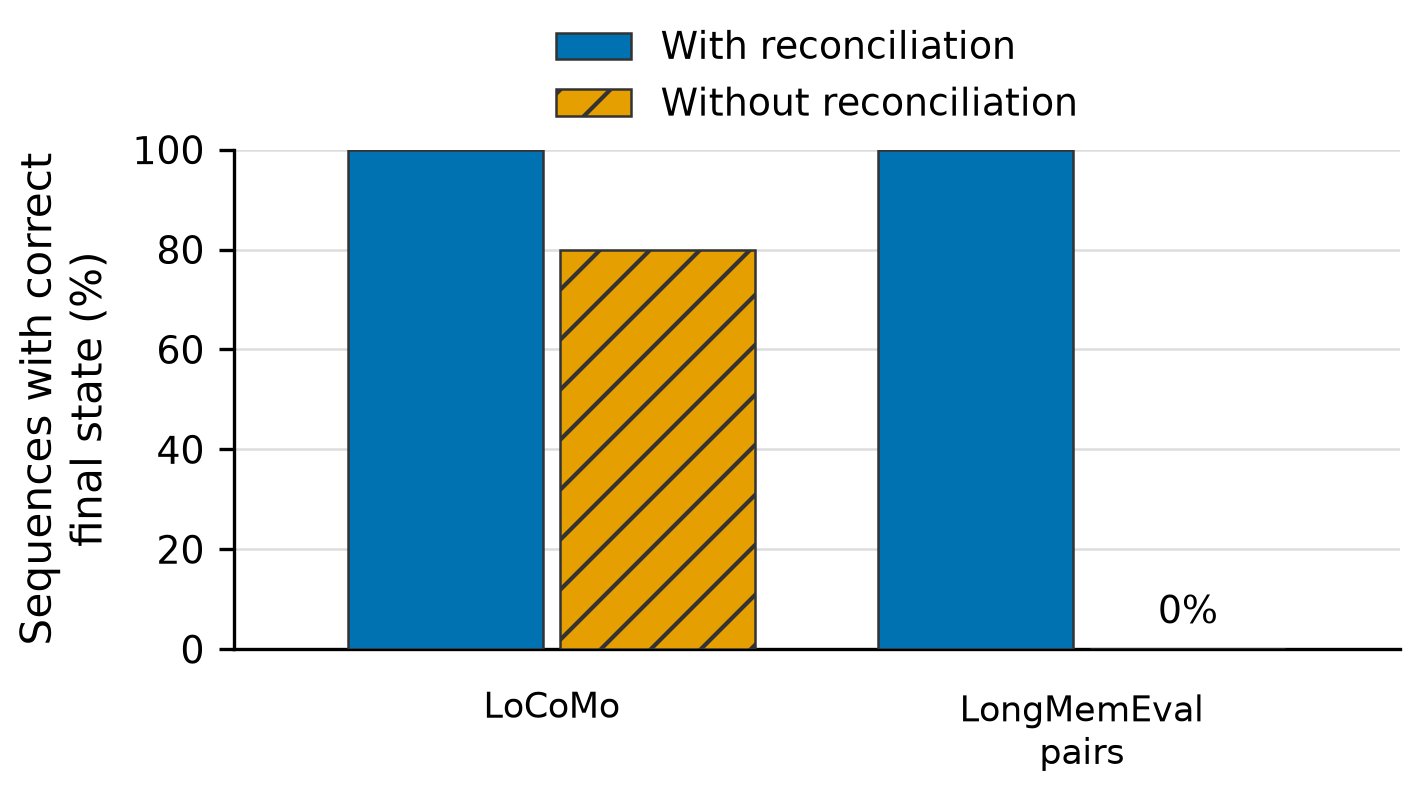}
\caption{Stashbird's final-state correctness with and without relationship reconciliation across ten annotated LoCoMo histories and LongMemEval evidence pairs. A sequence passes when all defined checks succeed after its final update.}
\label{fig_update_reconciliation}
\end{figure}

\noindent \textbf{Role of Update Reconciliation.}
We evaluate Stashbird through successive updates, restoring saved memory and checkpoints before processing each new batch. The experiment covers ten annotated LoCoMo histories processed session by session and ten selected LongMemEval pairs processed as an earlier message followed by a later correction. We compare updates with and without relationship reconciliation, which extends provenance for repeated facts and supersedes earlier relations after corrections. Both variants use supplied extraction outputs and the same reconciliation-decision rule, with all other update paths unchanged. Figure~\ref{fig_update_reconciliation} shows that Stashbird passes all final-state checks in all 20 sequences with reconciliation, compared with eight without it. Disabling reconciliation leaves four duplicate relations in LoCoMo and ten superseded relations active in LongMemEval. These results support Stashbird's incremental update correctness under the supplied semantic decisions.

\begin{figure*}[t]
\centering
\begin{minipage}{0.49\textwidth}
\centering
\includegraphics[width=\linewidth]{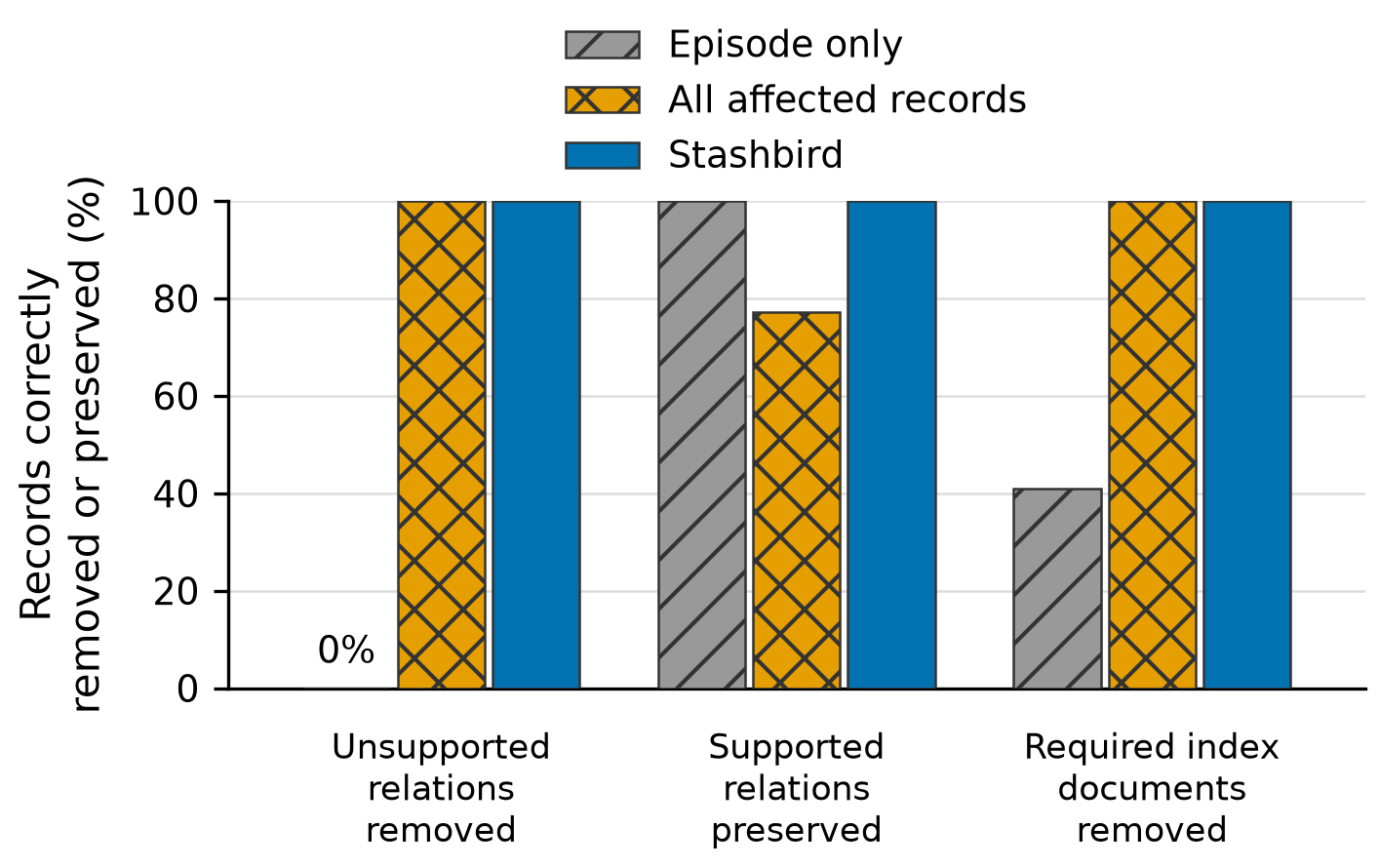}
\par\small (a) Removal and preservation on LoCoMo
\end{minipage}
\hfill
\begin{minipage}{0.49\textwidth}
\centering
\includegraphics[width=\linewidth]{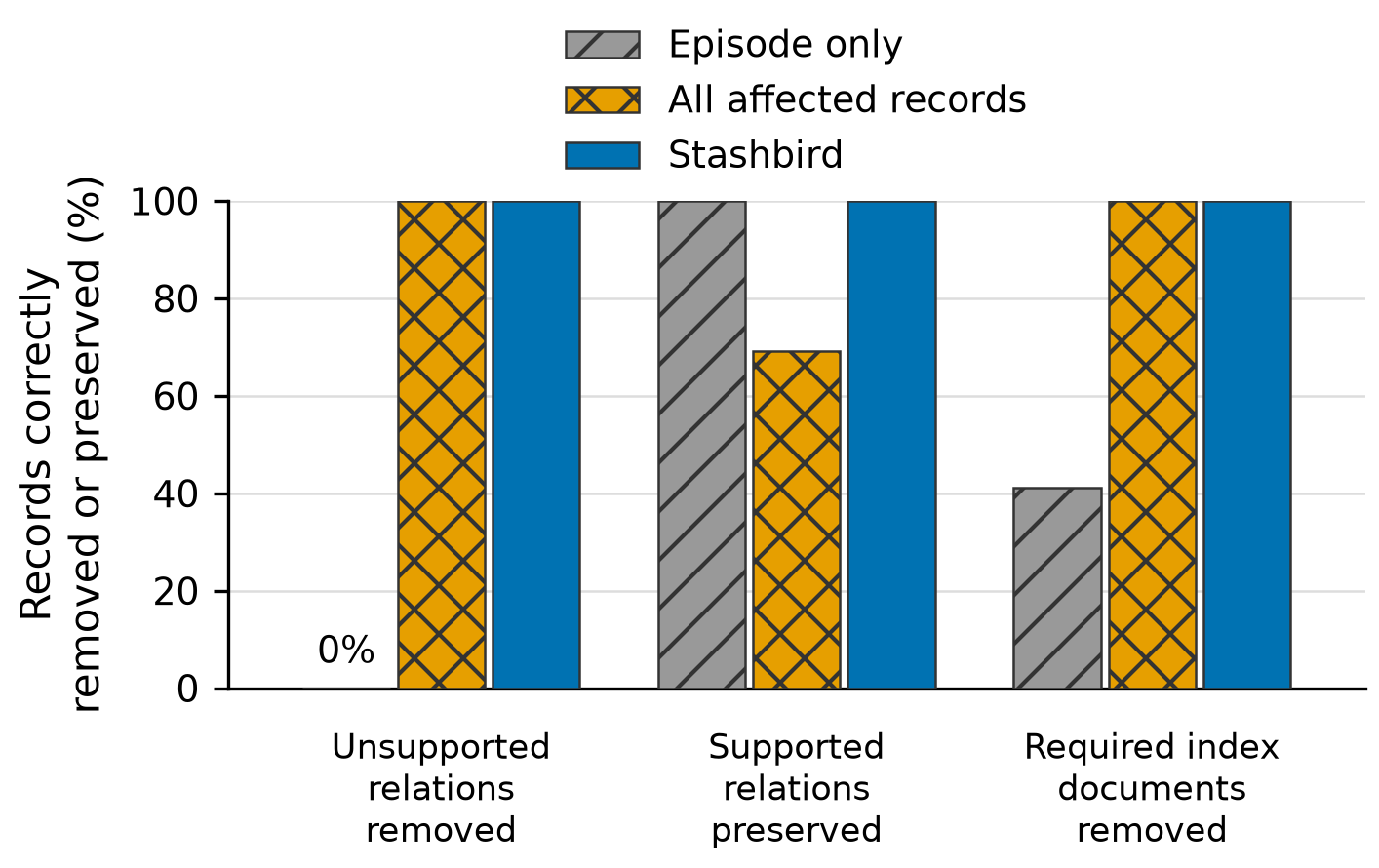}
\par\small (b) Removal and preservation on LongMemEval
\end{minipage}
\caption{Stashbird and comparison deletion policies. Bars show the percentage of eligible record occurrences satisfying each check. In axis order, denominators are 101, 1929, and 516 for LoCoMo and 50, 130, and 170 for LongMemEval. Records can recur across interventions. Property-list provenance is shown, with identical counts under MENTIONS-edge provenance.}
\label{fig_deletion_policy}
\end{figure*}

\noindent \textbf{Impact of Deletion Policy.}
We evaluate Algorithm~\ref{alg:deletion} across 203 interventions on ten LoCoMo histories and 60 interventions on ten selected LongMemEval histories, using annotation-defined facts and provenance. Each policy receives identical starting states and requests that remove one supporting episode, all annotated supports, or an episode outside the support sets. Episode-only deletion removes the episode and its indexed document. Deleting all affected records removes associated entities and relations regardless of remaining support. 

Figure~\ref{fig_deletion_policy} shows that Stashbird removes every unsupported relation and required index document while preserving every supported relation. Episode-only deletion leaves all unsupported relations behind, whereas deleting all affected records loses 441 supported relation occurrences in LoCoMo and 40 in LongMemEval. These comparisons demonstrate Stashbird's selective removal and preservation under the supplied provenance. Affected community documents are removed from the index while their graph summaries await regeneration.

\section{Discussion}
\label{sec:limitation}

We show that source-linked reconciliation maintains the required memory state during updates and deletion. Stashbird passes all 20 update sequences with reconciliation, compared with eight without it, and satisfies every removal and preservation check across 263 deletion interventions. The same mechanism therefore connects both lifecycle operations.

The benchmarks show a quality-efficiency tradeoff. Stashbird uses fewer prompt tokens than Hindsight, with higher accuracy on LongMemEval-S, comparable accuracy on EverMemBench, and lower accuracy on LoCoMo. Accuracy remains stable as the number of retrieved items changes. GroupMemBench results highlight strengths in temporal and user-implicit questions, while EverMemBench shows remaining challenges in evidence chaining and behavioral inference.

\noindent \textbf{Limitations.}
We did not sweep entity-matching thresholds or individually ablate the active retrieval components. QA evaluation uses benchmark-provided LLM judges without independent human validation. Efficiency measurements cover model calls and tokens on benchmark workloads, excluding database costs, storage, and end-to-end latency. Episode deletion leaves conversation summaries and rolling checkpoints unchanged, while affected community summaries remain stored pending regeneration.

\section{Related Work}
\label{sec:relatedwork}

Persistent agent-memory systems either manage textual history or build structured memory. MemGPT~\cite{packer2023memgpt} pages information between active context and archival storage, while MemoryBank~\cite{zhong2024memorybank} stores histories, summaries, and user portraits. Mem0~\cite{chhikara2025mem0} applies add, update, delete, or noop operations to extracted memories. Zep~\cite{rasmussen2025zep} links episodic, semantic, and community subgraphs while tracking temporal validity and contradictions. A-Mem~\cite{xu2025mem} evolves linked memory notes, Mnemis~\cite{tang2026mnemis} combines similarity retrieval with structured selection, and MemOS~\cite{li2025memos} and Hindsight~\cite{latimer2025hindsight} manage multiple forms of changing memory. iText2KG~\cite{lairgi2024itext2kg} incrementally constructs knowledge graphs while resolving duplicate entities and relations.

Graph-based retrieval methods include HippoRAG~\cite{gutierrez2024hipporag}, which uses Personalized PageRank over a knowledge graph, and G-RAG~\cite{dong2024don}, which reranks evidence using document connections and semantic graphs. Recent lifecycle work removes low-value graph nodes~\cite{rusu2026selective}, evaluates repeated memory updates and deletion~\cite{uddin2026recall}, or removes information from summaries, plans, and cached state through checkpoint restoration and replay~\cite{yao2026forgetting}.

\section{Conclusion}
\label{sec:conclusion}

We presented Stashbird, an episode-grounded memory system that uses source provenance to reconcile derived memory as evidence changes. Lifecycle studies show that reconciliation prevents duplicate and stale relations and removes unsupported state while preserving supported records and graph connectivity. Across four benchmarks, Stashbird achieves competitive question-answering accuracy with substantially lower prompt-token usage than reproduced Hindsight on the measured benchmarks.

\section*{Ethical Considerations}

This work aims to improve the consistency and efficiency of conversational agent memory. Persistent storage of personal and group conversations raises privacy and consent concerns, while incorrect or misattributed memories may affect subsequent answers. Deployment should respect participant permissions and retention preferences and communicate the scope of deletion.



\FloatBarrier
\bibliographystyle{unsrt}
\bibliography{reference}


\appendix




\newpage

\end{document}